\documentclass[letterpaper, 10 pt, conference]{ieeeconf}  

\IEEEoverridecommandlockouts                 

\usepackage{multibib}
\usepackage[utf8]{inputenc}

\usepackage{algpseudocode}           
\usepackage{algorithm}
\usepackage{amsmath}
\usepackage{amssymb}
\usepackage{bbm}
\usepackage{mathtools}
\usepackage{commath}
\DeclareMathOperator*{\argmaxA}{arg\,max}

\usepackage{url} 

\usepackage{multirow}                
\usepackage{float}
\usepackage{soul,booktabs}

\title{\LARGE \bf
Leak Event Identification in Water Systems Using High Order CRF
}

\author{Qing Han, Wentao Zhu and Yang Shi\\
}

\begin{document}

\maketitle
\thispagestyle{empty}
\pagestyle{empty}

\begin{abstract}

Today, detection of anomalous events in civil infrastructures (e.g. water pipe breaks and leaks) is time consuming and often takes hours or days. Pipe breakage as one of the most frequent types of failure of water networks often causes community disruptions ranging from temporary interruptions in services to extended loss of business and relocation of residents. In this project, we design and implement a two-phase approach for leak event identification, which leverages dynamic data from multiple information sources including IoT sensing data (pressure values and/or flow rates), geophysical data (water systems), and human inputs (tweets posted on Twitter). In the approach, a high order Conditional Random Field (CRF) is constructed that enforces predictions based on IoT observations consistent with human inputs to improve the performance of event identifications. 

Considering the physical water network as a graph, a CRF model is built and learned by the Structured Support Vector Machine (SSVM) using node features such as water pressure and flow rate. After that, we built the high order CRF system by enforcing twitter leakage detection information. An optimal inference algorithm is proposed for the adapted high order CRF model. Experimental results show the effectiveness of our system.

\end{abstract}

\section{INTRODUCTION}

Water is a critical resource and a lifeline service to communities worldwide; they are essential for sustaining the economic and social viability of a community. Often the infrastructures that capture, deliver and store water in cities and communities are many decades old; with the rise in urban populations, these infrastructures have become more increasingly complex and vulnerable to failures due to natural, technological and manmade events.  For example, pipe breakage is one of the most frequent types of failure of water networks and often causes community disruptions. Based on a report from Los Angeles Department of Water and Power (LADWP), Los Angeles (LA) has been experiencing an unusual increase in pipe beaks and leaks, mainly in old pipes that are susceptible to corrosion problems and pipe joint displacements caused by surface deformations. Pipe bursts may also cause transportation network collapse and water loss often lead to additional energy expenditures for transporting water from natural resources to the end users. Extreme weather and rainfall events (e.g. Hurricane Sandy, El Ni\~{n}o 2016) stresses already weakened pipes to the point of causing major pipe breaks and significant increases in leak rates and thus major pipe breaks of failures. It represents a very high cost vulnerability and is associated with public health implications and wastage of a limited resources.

Pipe breaks or bursts often reduce pressure heads and increase flow rates at failure point. IoT sensing data from water infrastructures can track the changes of the network in a timely manner, and reflect a certain level of failures in the network. However, these measurements are limited by ($a$) sensor locations (static sensors), ($b$) the number of sensors installed (due to high cost), and ($c$) they are highly correlated with each other. Therefore, it is hard to isolate the damaged pipes by these data itself, but aggregating with external sources will be helpful. Human reports related to leak events may complement the limitations of IoT observations. Because human sensing are more dynamic, reliable, and accessible. The key of combination/fusion is that information from different data sources indicate the presence of a problem at different 

Conditional Random Field (CRF) \cite{lafferty2001conditional} has been successfully used in structured prediction problems in the undirected graphic models. The main advantage of CRF is that it tries to model $p(\mathbf{y}|\mathbf{x})$ instead of $p(\mathbf{x}, \mathbf{y})$ given the observation $\mathbf{x}$. However, the CRF model only allows the adjacent relationship of $\mathbf{y}$ due to the markov property of CRF. The high order CRF model \cite{ye2009conditional} extends it to exploit high-order dependencies, which provides substantial performance improvement. In this paper, we first use the CRF to model the leak event in the water system. Then a high-order CRF model is used by enforcing human inputs.

The rest of this paper is organized as follows. Related work on leak detection, CRF and Structured Support Vector Machine (SSVM) are introduced in Section \ref{related}. We describe the proposed approach in Section \ref{approach} that is extensively validated using data from the hydraulic simulator in Section \ref{validation}. Section \ref{conclusion} concludes the paper.   

\section{RELATED WORK}\label{related}

There has been substantial work on single leak detection. The primary method to determine a leak event is based on the measurement of leak-related vibro-acoustic phenomena with the help of expensive and sophisticated devices whose efficiency largely depends on the operator skills \cite{hessel1999neutral,rajtar1997pipeline}. Machine learning techniques have been suggested for leak identification problems, such as maximum likelihood methods \cite{poulakis2003leakage,rougier2005probabilistic}, SCEM-UA algorithm \cite{puust2006probabilistic}, and neural network \cite{ai2006pipeline,mashford2009approach}. The results, however, are not solid because these approaches were evaluated by specific use cases that are not sufficient to address the general performance.

Weng et. al.~\cite{weng2013graphical} developed a fast on-line state estimation in electrical system. Belief propagation and variational belief propagation are applied to the proposed graphical model. The gain of using these methods is not only in computation accuracy but also in computation time. Beyond these, the estimation algorithm can scale up well. However, the model is only for electrical pressure estimation from the noisy observation, and can not be integrated with the outside information directly. 

Ahmed et. al.~\cite{qa2012icic} considered a gasoline leakage problem. By defining different stages of the system, i.e. healthy, minor fault and faulty stages. A hidden markov model is used in the system. However, they didn't attempt to build the estimation for the leakage location, and the model cannot be applied to undirected or direction changeable problems, such as water systems.

Recently, high order conditional random field models are emerging \cite{kohli2009robust,wegner2013higher}, including deep structural model \cite{zhu2016adversarial}. These models enforce label consistency in the CRF model and obtain better performance than previous models. We adapted the high order CRF model to the leakage detection in the water system. Based on the special features we have in the waterpipe network, we proposed an optimal solution for the problem.

\section{Two-phase Approach to Leak Event Identification}\label{approach}

Pipe leaks or bursts often lead to changes in pressure heads and flow rates, which can be used to obtain critical information on which parts of the system are suffering the effects of water pipe failures. However, the IoT sensing data itself (pressure and flow rate) may not enough to locate all leak events due to (\textit{a}) limited observations (inaccessible locations and high cost) and (\textit{b}) highly correlated features (densely connected network). In real world, the damage to underground infrastructures is often hidden, and most pipe failures are silent until they are noticed by people. Thus, human inputs are integrated in the inference process to complement limitations of the IoT observations. 

\subsection{Problem Formulation}

A water system is modeled as an undirected graph $G(\mathcal{V}, E)$ (water can flow in both directions) with vertices $\mathcal{V}$ that represent the nodes (joint of pipes), and edges $E$ that represent pipelines. $|\mathcal{V}|$ equals to the number of nodes in the network. A set of pressure and/or flow rate sensors $\mathcal{A}$ are simulated using the hydraulic simulator. $\mathcal{C}$ is a set of subsets which contains relevant vertices from human sensing. We consider $X$ as a set of observations, i.e. the measurements of pressure values and/or flow rates collected from sensors, and $Y$ as a set of event variables, i.e. the leakage states (leaking or not) of each node that we wish to identify. An arbitrary assignment to $X$ is denoted by a vector $\mathbf{x} = \{x_{a}: a \in \mathcal{A}\}$. Similarly for $Y$, an assignment $\mathbf{y} = \{y_{v}: v \in \mathcal{V}\}$ is a vector of labels taking from the label set $\mathcal{L} = \{0, 1\}$ where $y_{v} = 1$ indicates a leak at location $v$. Note that the leak event is assumed to occur at the joint of pipes for simplicity.
\subsection{Two-Phase Approach}
In the first part, We will discuss how to build and learn the CRF model. After that, we will explore the high order CRF model by adding human report in the built CRF model. At last, a greedy inference method is proposed for the high order CRF model. 


%



\subsubsection{\textbf{Phase I: Learning IoT Observations}} \label{IOT}

Conditional Random Field (CRF) is an undirected graphical model that has been widely used for structured prediction \cite{lafferty2001conditional}. Based on the formulation, ($X$, $Y$) is a conditional random field because the random variable $Y$ is indexed by the vertices of $G$ and $y_{v}$ conditioned on $\mathbf{x}$ obey markov property with respect to the graph $G$: $p(y_{v} | \mathbf{x}, y_{v'}, v' \neq v) = p(y_{v} | \mathbf{x}, y_{v'}, (v, v') \in E)$ \cite{wiki:crf}. The conditional distribution $p(\mathbf{y} | \mathbf{x})$ can then be modeled and trained by using machine learning based techniques. 

Structured Support Vector Machine (SSVM) can be used for the learning and inference of CRFs by generalizing the Support Vector Machine (SVM) classifier to do structured learning. 

 \begin{equation}
 \begin{split}
\min\limits_{\mathbf{w}} & \sum_{n=1}^{N} \max_{\mathbf{y}_{n} \in Y} (\Delta(\hat{\mathbf{y}}_n, \mathbf{y}_{n}) + \mathbf{w}^T \theta(\hat{\mathbf{y}}_n, \mathbf{x}_{n}) - \mathbf{w}^T \theta(\mathbf{y}_n, \mathbf{x}_n)) \\ &+ \frac{C}{2} \norm{\mathbf{w}}^{2}
\end{split}
\end{equation}
where $\Delta(\hat{\mathbf{y}}_v, \mathbf{y}_{v}) = \sum_{v \in \mathcal{V}} \mathbbm{1}[\hat{\mathbf{y}}_v \neq \mathbf{y}_{v}]$ is the hamming loss between the prediction $\mathbf{y}_{v}$ and leakage observation $\hat{\mathbf{y}}_v$, $C$ is the cost factor related to the trade off between the empirical loss and regularization, $N$ is the number of training samples.  

In the inference period, we use a structured linear predictor based on the learned SSVM model: 

\begin{equation} \label{crfinfer}
\hat{\mathbf{y}} = \argmaxA_{\mathbf{y} \in Y}  \mathbf{w}^{T}\theta( \mathbf{x}, \mathbf{y})
\end{equation}
%
where $\hat{\mathbf{y}}$ is a vector of predicted structured labels, $\mathbf{w}$ are parameters that are learned from data, and $\theta$ is defined by the user-specified structure of the model \cite{muller2013pystruct}. To compute the $argmax$, several inference solvers, e.g. Quadratic Pseudo-Boolean Optimization (QPBO) and Alternating Directions Dual Decomposition (AD3), can be applied. 

The inference outcome of the model learned on IoT observations is $\mathcal{S} = \{v : \hat{y}_{v} = 1 \wedge v \in \mathcal{V}\}$, representing a subset of $\mathcal{V}$ which are predicted as leaking positions. This set will be updated after Phase II. Notice that, we can also obtain the label assignment probability in this model \cite{platt1999probabilistic}.

%
%
  
\subsubsection{\textbf{Phase II: High order CRF with Human Inputs}}  
\label{hoc}

It is natural to think that there are higher possibilities for one subarea to have pipeline break if some human living around reported in social networks. To leverage human inputs, we bring in social media, Online Social Network (OSN), to incorporate human sensing. OSN has become a major platform for information sharing in which we can mine interested patterns \cite{kumar2014twitter}. We apply the Tweet Acquisition System (TAS) developed at UCI to selectively collect tweets relevant to leak events from Twitter, and use the associated geographic information to track and locate the risky area. The human input, however, is unable to specify the exact position of the damage due to various social behaviors. Therefore, it is considered as high order potentials \cite{kohli2009robust} in the inference process to enforce event consistency. We assume that twitter information can reflect true events with high confidence. That is, based on the content, location, and the number of tweets we can locate the faulty region at different levels of granularity.  

Let $\mathcal{C} = \{c : c = \{v : |l_{c} - l_{v}| < \gamma \wedge v \in \mathcal{V}\}\}$ represent a set of subsets of $\mathcal{V}$ (i.e. a set of cliques) infered from human inputs. Here, $|l_{c} - l_{v}| < \gamma$ indicates that nodes $v$ whose distance to the location of clique $c$ ($l_{c}$) identified by the GeoTag of tweets is less than the threshold $\gamma$. That is, nodes within a certain distance from the location of the tweet are likely to leak. The high order potential $\Phi_{c} : \mathcal{L}^{|c|} \rightarrow \mathbb{R}$ is defined over this clique assigns a cost to each possible configurations (or labelings) of $\mathbf{y}$. Because we assume that the effects of human inputs on leak event identification is non-negative, we have

\begin{equation}\label{highOrder}
\Phi_{c} = 
			\left\{
                \begin{array}{ll}
                  0 \hspace{9mm} if \hspace{2mm} \exists v \in \mathcal{S} \hspace{2mm} for \hspace{2mm} v \in c\\
                  Inf \hspace{5mm} o.w.\\
                \end{array}
             \right.
\end{equation}
where $S$ is the leakage set obtained by the above CRF model. 

The high-order CRF model enforcing the human input can be obtained as 
\begin{equation}\label{energy2}
\min  \sum\limits_{v \in \mathcal{V}} \Phi_{v}(y_{v}) + \sum_{\mathclap{\substack{(v, v') \in E}}} \Phi_{v, v'}(y_{v}, y_{v'}) + \Phi_{c}
\end{equation}
%
where the unary potential $\Phi_{v}(y_{v})$ is defined as the minus entropy (corresponding to maximum entropy) of $y_v$ if the vertex $v$ is in the human reported leaking cliques and all the vertices in that clique are not in the set $S$. If the vertex $v$ is not in the human reported leaking cliques or a vertex in the clique is in the $S$, $\Phi_{v}(y_{v})$ can be defined as $- \mathbf{w}^{T}\theta( \mathbf{x}, \mathbf{y})$, which means the prediction for the vertex $v$ is the CRF inference result as (\ref{crfinfer}) since there is no extra information adding into the system. Because the leakage events among the vertices of $G$ are conditionally independent given the observations $X$, the pairwise terms $\Phi_{v, v'}(y_{v}, y_{v'})$, in our case, are constant. 


\subsubsection{\textbf{Inference}}  
\label{infer}

According to (\ref{highOrder}), an event inconsistency can push the energy to the infinity. Therefore, Algorithm 1, a greedy inference algorithm, is to update $\mathcal{S}$ by adding a node $v^{*}$ from the clique with $\Phi_{c} = Inf$ into $S$ if $v^{*} = \argmaxA_{v \in c} H(y_{v})$ (\ref{entro}) and $H(y_{v^{*}}) > \Gamma$. 
%
\begin{equation}\label{entro}
	H(y_{v}) = - \sum\limits_{i=0}^{1} p_i(\hat{y}_{v})\log p_i(\hat{y}_{v})
\end{equation}
Note that $p_i(\cdot)$ can be obatined by machine learning based techniques applied in Section \ref{IOT}. In this manner, Algorithm 1 can remove the inconsistency between IoT observations and human inputs to minimize the energy function given in (\ref{energy2}). 

\begin{algorithm}
  \caption{Greedy algorithm for the integration of human inputs (the high order potential)}
  
  \begin{algorithmic}[1] 
  	\State \textbf{Input} $\mathcal{S}$, $\mathcal{C}$
    \State \textbf{Output} \textit{updated}($\mathcal{S}$)
    \State \textbf{Objective} $\min_{\mathcal{C}} E[\mathbf{y}]$
    \item[]
    \State $\mathcal{S} = \{v : \hat{y}_{v} = 1 \wedge v \in \mathcal{V}\}$
    \State $\mathcal{C} = \{c : c = \{v : |l_{c} - l_{v}| < \gamma \wedge v \in \mathcal{V}\}\}$
    
    \For {$c$ in $\mathcal{C}$}
    	
        \If{ $\Phi_{c} =$ 0}
        	\State continue 
        \Else{}              
        	\State $v^{*} = \argmaxA_{v \in c}$ H($y_{v}$)  
            \If{ $H(y_{v^{*}}) > \Gamma$ }
            	\State $\mathcal{S} = \mathcal{S} \cup \{v^*\}$
            \EndIf
        \EndIf
    \EndFor
    
  \end{algorithmic}
\end{algorithm}

\section{Experiment and Results}\label{validation}

In this section, we begin by presenting the datasets and parameters under which the simulations are conducted, and describe the implementation of the high order CRF model. 

\subsection{Datasets and Experiment Setup}

\begin{figure}
\centering 
\includegraphics[height=2.3in, width=3in]{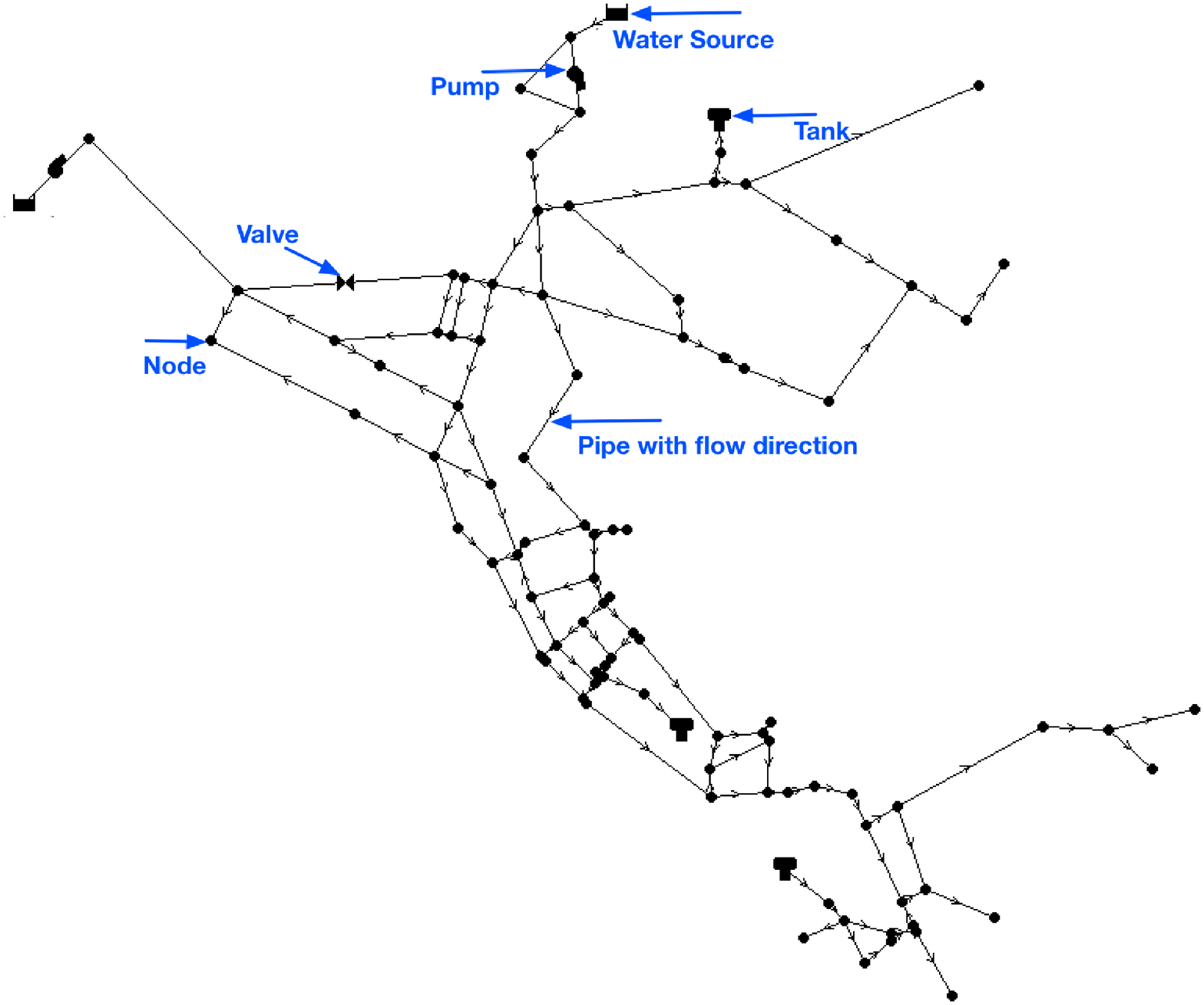}
\caption{A water network provided by EPANET with 118 pipes, 96 nodes, 2 pumps, a valve, 
3 tanks, and 2 water sources.} \label{map}
\end{figure}

The IoT sensing data is generated using a commercial grade hydraulic simulator EPANET \cite{rossman2000epanet}. Figure \ref{map} shows a real-world based water network provided by EPANET. 
The elevations of pipes varies with the topography, and each pipe has four attributes - length, diameter, roughness coefficient, and status (open or close controlled by valve). Each node has a pattern of time variation of the demand (i.e. consumption). A leak event is assumed to occur at nodes, since the interconnect points are more risky than others. We assume a fully IoT observations composed of pressure values and flow rates with the number of sensors $|\mathcal{A}| = 218$. That is, each data sample has $218$ features. 

Extensive simulations are run on EPANET to generate sufficient observations for training. The number of training sets and testing sets are $10,000$ and $2,000$ separately. For each simulation run, different numbers of leak event(s) are generated randomly with different locations, sizes, and starting time. The sampling frequency of the sensors is 15 minutes. 

\subsection{Implementation}

In Phase \textbf{I}, the IoT observations are trained using SSVM with penalty parameter $C = 0.25$. We use MATLAB with libsvm packet \cite{CC01a} to first predict possible leak event(s) with the leak probability that will be used in the inference process in Phase \textbf{II}. 

As mentioned in Section~\ref{hoc}, tweets posted on Twitter are used for social media dataset that can complement the limitations of IoT observations. We simulate the human inputs by assigning a probability $p$ that people will report a leak around the true event position. The confidence of the human reports can be simulated by changing the threshold $\gamma$. To avoid the inconsistency between the IoT observations and human inputs, the predicted label of the node with the highest entropy that is larger than $\Gamma$ will be set to $1$ as running Algorithm 1. 


\subsection{Result}

We use Hamming score as the evaluation for the results. Hamming score is defined as $\frac{P \cap T}{P \cup T}$ in our case, where $P$ is a set of locations predicted as the leak spots, and $T$ is a set of true event locations. The score is bounded by $1$, and the higher the score the better the performance.

In previous work, we applied random forest on this dataset. The hamming score without human inference is \textbf{0.65} and the one with high order potential is \textbf{0.74}. In this paper, the best hamming score we have by learning the model and inference using SSVM without the integration of high order potential is $\textbf{0.9566}$. Table~\ref{result} shows that the hamming score varies with the parameters after we consider human sensing. Clearly, the integration of high order potential improves the performance. With more accurate human reports, the performance is even better.


\begin{table}[t]
 \newcommand{\tabincell}[2]{\begin{tabular}{@{}#1@{}}#2\end{tabular}}
 \centering \footnotesize
 \caption{\label{result} Hamming score of leak event identification with learning using SSVM and inference with high order potential.}
 \begin{tabular}{@{}cccc@{}}
 \toprule
 \textbf{$p$} & \multicolumn{1}{c}{\textbf{$\gamma$ (unit: meter)}} &  \multicolumn{1}{c}{$\Gamma$} &  \multicolumn{1}{c}{\textbf{Hamming Score}} \\
 \midrule
	$0.3$ & \tabincell{l}{$2$} & \tabincell{l}{$0$} & \tabincell{l}{$\mathbf{0.9654}$} \\
	$0.3$ & \tabincell{l}{$3$} & \tabincell{l}{$0.04$} & \tabincell{l}{$\mathbf{0.9622}$} \\
	$0.7$ & \tabincell{l}{$2$} & \tabincell{l}{$0$} & \tabincell{l}{$\mathbf{0.9789}$} \\
	$0.7$ & \tabincell{l}{$3$} & \tabincell{l}{$0.04$} & \tabincell{l}{$\mathbf{0.9701}$}  \\
 \bottomrule											
 \end{tabular}
\end{table}

 

\section{conclusions} \label{conclusion}
In the project, we constructed a CRF model to model the structure leak detection in the water system. The structured SVM is used for the built CRF model. To fully take advantage of human inputs, a high order CRF model is adapted to our problem. An optimal greedy inference algorithm is proposed for the high order CRF model in our case. Experimental results show the effectiveness of the proposed system and the desirable detection performance.

\addtolength{\textheight}{-12cm}  

\bibliographystyle{IEEEtran}
\bibliography{bibliography}

\end{document}